\documentclass{article}
\usepackage{spconf,amsmath,graphicx}
\usepackage{amssymb,amsfonts}
\usepackage{mathrsfs}
\usepackage{multirow}
\usepackage{booktabs}
\usepackage[colorlinks=true, 
allcolors=blue,
bookmarksopen=true,
bookmarksnumbered=true]{hyperref}
\usepackage[table,xcdraw]{xcolor}
\usepackage{stmaryrd}
\usepackage{enumitem}

\title{Towards Trustworthy Multi-label Sewer Defect Classification via Evidential Deep Learning}
\name{Chenyang Zhao$^1$, Chuanfei Hu$^1$, Hang Shao$^2$, Zhe Wang$^3$, Yongxiong Wang$^3$}
\address{
	\small$^1$Key Laboratory of Measurement and Control of CSE Ministry of Education, School of Automation, Southeast University, Nanjing, China \\
	\small $^2$School of Computer Science and Engineering, Nanjing University of Science and Technology, Nanjing, China \\
	\small $^3$ School of Optical-Electrical and Computer Engineering, University of Shanghai for Science and Technology, Shanghai, China \\
	\small \{zhaocy, cfhu\}@seu.edu.cn, shaohang@njust.edu.cn, 201440049@st.usst.edu.cn, wyxiong@usst.edu.cn
	\thanks{Chenyang Zhao and Chuanfei Hu contributed equally to this work. Corresponding author: Chuanfei Hu (cfhu@seu.edu.cn).}
}
\begin{document}
%
\maketitle
\begin{abstract}
	
An automatic vision-based sewer inspection plays a key role of sewage system in a modern city. 
Recent advances focus on utilizing deep learning model to realize the sewer inspection system,
benefiting from the capability of data-driven feature representation.
However, the inherent uncertainty of sewer defects is ignored,
resulting in the missed detection of serious unknown sewer defect categories.
In this paper, we propose a trustworthy multi-label sewer defect classification (TMSDC) method, 
which can quantify the uncertainty of sewer defect prediction via evidential deep learning.
Meanwhile, a novel expert base rate assignment (EBRA) is proposed to introduce the expert knowledge for describing reliable evidences in practical situations.
Experimental results demonstrate the effectiveness of TMSDC and the superior capability of uncertainty estimation is achieved on the latest public benchmark. 

\end{abstract}
\section{Introduction}
\label{sec:intro}


Underground sewage system is one of the most vital lifelines in a modern city \cite{Nikola2014HAZard},
which can guarantee the community health, safety, and manufacture.
Vision-based inspection method is widely applied to maintain the underground sewage system \cite{Liu2013State}.
The internal situations across the sewer pipes can be captured via a remote mobile vehicle, while the sewer inspectors diagnose the defects with a long time of looking at a screen.
Such manual inspection is not only laborsome and time-consuming, 
but also may cause ophthalmic diseases during the high-frequency illumination of the screen. 
Consequently, how to construct an automatic sewer inspection method has long been a research topic attracting constant attention in the field of sewer inspection \cite{Rayhana2021Automated}.

Recently, deep learning model 
has received substantial interest in industrial applications \cite{Hu2020Efficient, Su2022PVEL}.
In the vision-based sewer inspection community, deep learning also attracts increasing attention from both academia and industry 
\cite{Chen2018Intelligent, Xie2019Automatic, Haurum2021Sewer}. 
Here, we focus on the sewer defect classification in the setting of multi-label, 
in which multiply defect classes in an image are recognized simultaneously.
Although these deep learning-based methods have achieved acceptable performances of sewer defect classification, 
while the inherent uncertainty of sewer defects might not be considered sufficiently in real-world applications \cite{ROGHANI2019Dealing}.
For instance, some categories of sewer defects are not appeared from historical data, in sense that, the trained sewer defect classification model has not seen these \emph{unknown} defects which is the samples out of knowledge.
Existing deep learning-based methods \cite{Chen2018Intelligent, Xie2019Automatic, Dang2022DefectTR} for sewer defect classification could not describe the magnitude of epistemic uncertainty across \emph{known} and \emph{unknown} sewer defect categories. 
The model would be over-confident to ``trust'' the prediction, 
resulting in the missed detection of serious unknown sewer defect categories.         

In this paper, we propose a trustworthy multi-label sewer defect classification (TMSDC) method for unknown sewer samples setting.
To enable the multi-label sewer defect classification model to ``\emph{know unknown}'', 
we cast the task as an uncertainty estimation problem via evidential deep learning (EDL) \cite{sensoy2018evidential}.
EDL describes the uncertainty via a Dirichlet distribution of class probability,
which can be seen as an evidence collection process via a deep neural network.
The collected evidence is leveraged to quantify the uncertainty of sewer defect prediction,
for instance, \emph{unknown} sewer defect would present a high uncertainty explicitly.
%
Moreover, we introduce the expert knowledge to model the uncertainty and propose an expert base rate assignment (EBRA), 
in which the realistic base rate can provide reliable diagnosis of sewer defects in practical situations \cite{josang2016subjective}. 
It is noteworthy that TMSDC can quantify the uncertainty effectively of model whose capability of distinguishing the known categories would only be weakened slightly.
The main contributions are summarized as follows:

\begin{itemize}
	\item A trustworthy multi-label sewer defect classification (TMSDC) method is proposed, 
	which can quantify the uncertainty of sewer defect prediction via evidential deep learning (EDL).
	To the best of our knowledge, we are among the first to introduce EDL for promoting the trustworthiness of sewer inspection.
	\item A novel expert base rate assignment (EBRA) is proposed to model reliable evidences in practical situations via expert knowledge.
	\item The effectiveness of TMSDC is demonstrated with diverse metrics on Sewer-ML \cite{Haurum2021Sewer} which is a large-scale public benchmark in the field of sewer inspection.
\end{itemize}

\begin{figure}[!t]
	\centering
	\includegraphics[width=\linewidth]{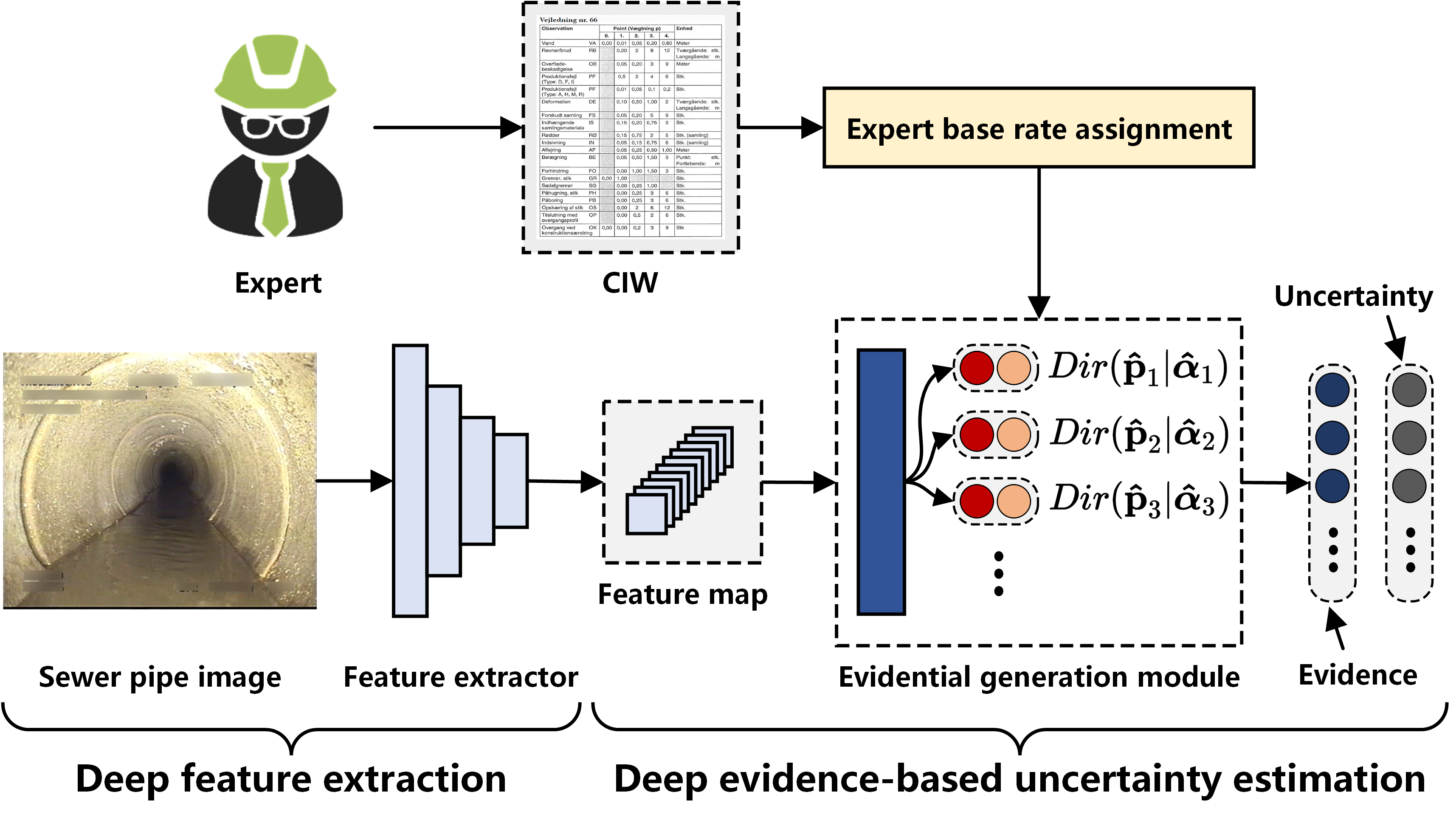} 
	\caption{The overall framework of TMSDC. 
	}
	\label{fig:framework}
\end{figure}

\vspace{-4mm}
\section{Methodology}
\label{sec:method}

TMSDC aims to quantify the uncertainty of sewer defect predictions via EDL in a deep learning-based paradigm,
ensuring the reliable performance of multi-label sewer defect classification.
The overall framework of our proposed method consists of deep feature extraction and deep evidence-based uncertainty estimation, as illustrated in Fig.~\ref{fig:framework}. 
Specifically, given a sewer pipe image $\mathbf{x}$ whose feature map $\mathbf{f} \in \mathbb{R}^{C \times H \times W}$ is first obtained via a deep learning-based feature extractor $F$ as follows:
\vspace{-1mm}
\begin{equation}
	\mathbf{f} = F(\mathbf{x}| \theta_{F}),
\vspace{-1mm}
\end{equation}
where $\theta_{F}$ denotes the learnable parameters of feature extractor, $C$ is the number of channels and $H \times W$ is the resolution.
Then, an evidential generation module is modeled to collect the deep evidence of sewer defect class, 
in which the sewer defect probability is assumed to follow a Dirichlet distribution.
Formally, the evidence $\mathbf{{e}}_K$ and uncertainty $\hat{\mathbf{{u}}}_K$ of $K$ sewer defect classes can be formulated via the evidential generation module (EGM) $G$ as follows:
\vspace{-1mm}
\begin{equation}
	\{ \mathbf{{e}}_K, \hat{\mathbf{{u}}}_K \} = G(\mathbf{f}|\theta_{G},Dir(\hat{\mathbf{{p}}_{K}} | \hat{\boldsymbol{\alpha}}_{K}), \hat{\omega}),
\vspace{-1mm}
\end{equation}
where $\theta_{G}$ presents the learnable parameters of the EGM.
$Dir(\hat{\mathbf{{p}}_{K}} | \hat{\boldsymbol{\alpha}}_{K})$ is a Dirichlet distribution, 
$\hat{\mathbf{{p}}_{K}}$ and $\hat{\boldsymbol{\alpha}}_{K}$ are the predicted probability and the Dirichlet parameter, respectively. 
$\hat{\omega}$ denotes a subjective opinion adjusted via EBRA.


\subsection{Deep evidence-based uncertainty}\label{sec:edl}
\textbf{Evidential Deep Learning.} Recent EDL \cite{sensoy2018evidential} is widely utilized to estimate the uncertainty of classification results \cite{Han2022Trusted, zhao2022seed}, 
introducing the evidence framework of Subjective Logic (SL) \cite{josang2016subjective}.
SL defines a subjective opinion by explicitly considering the dimension of uncertainty derived from evidence vacuity.
Here, we first describe the original EDL for classification with $K$ categories.
A subjective opinion for a sample with $K$-dimensional class domain can be first formulated as a triplet $\omega = (\mathbf{b}, u, \mathbf{a})$ 
consisting of the belief mass $\mathbf{b} = \{b_{1}, b_{2}, \dots, b_{K}\}$,  the uncertainty $u$, and the base rate distribution $\mathbf{a} = \{a_{1}, a_{2}, \dots, a_{K}\}$.
For the $k$-th dimensional class, the probability mass $p_k$ of the subjective opinion $\omega$ is defined as:
\vspace{-1mm}
\begin{equation}\label{equ:pro_opinion}
	p_k = b_k + a_{k}u.
\vspace{-1mm}
\end{equation}
Since $p_k$ should be treated as a probability, the entities of $\omega$ are constrained by $u+\sum_{k=1}^K b_k=1$ and $\sum_{k=1}^K a_{k} =1$.
Formally, the probability mass $\mathbf{{p}}$ of $\omega$ follows a Dirichlet distribution with parameter $\boldsymbol{{\alpha}} = \{ \alpha_{1}, \alpha_{2}, \dots, \alpha_{K} \}$:
\vspace{-1mm}
\begin{equation}
	Dir(\mathbf{p} | \boldsymbol{\alpha})= \left\{
	\begin{aligned}
		&\frac{1}{B(\boldsymbol{\alpha})} \prod_{k=1}^K p_k^{\alpha_k-1} & \text { for } \mathbf{p} \in \mathcal{S}_K \\ 
		&0  &\text { otherwise }
	\end{aligned}
\right. ,
\vspace{-1mm}
\end{equation}
where $B$ is the Beta function, and $\mathcal{S}_K  = \{ \mathbf{p} | \sum_{k=1}^{K} p_k = 1  \text{ and } p_k \in [0,1],  \forall k\}$ is the $K$-dimensional unit simplex.
Then, the evidence $\mathbf{e} = \{e_{1}, e_{2}, \dots, e_{K}\}$ of $\omega$ are linked with Dirichlet parameter $\boldsymbol{{\alpha}}$ based on DST as follows:
\vspace{-1mm}
\begin{equation}
	\alpha_{k} = e_{k} + a_{k}W,
\vspace{-1mm}
\end{equation}
where $e_{k} \in  [0, +\infty)$ obtained directly from the last layer of neural network with a non-negative activation function, such as Rectified Linear Units (ReLU).
$W$ is the weight of uncertain evidence set as $K$ empirically.
Following the Dirichlet assumption, the expectation of $\mathbf{{p}}$ is given by:
\vspace{-1mm}
\begin{equation}\label{equ:pro_dir}
	\mathbb{E}\left(p_k\right)=\frac{\alpha_k}{\sum_{k=1}^K \alpha_k}=\frac{e_k+a_k W}{W+\sum_{k=1}^K e_k}.
\vspace{-1mm}
\end{equation}
When $a_{k}$ is set $1/K$, the Dirichlet parameter $\alpha_{k}$ can be formulated as $\alpha_{k} = e_{k} + 1$,
in which $u$ and $b_k$ can be determined by the parameter as
$
	u =K / \sum^{K}_{k=1}\alpha_{k}  \ \text{ and }  \ b_k = (\alpha_{k}-1) / \sum^{K}_{k=1}\alpha_{k}.
$
Thus, the probability of sample with evidence $e_k$ for $k$-th class can be predicted by Eq.~\ref{equ:pro_opinion} (or Eq.~\ref{equ:pro_dir}) simultaneously.

\textbf{EDL for multi-label sewer defect classification.} 
Since the $K$-dimensional predicted probability $\mathbf{{p}}$ of the popular multi-label classifier may not belong to $\mathcal{S}_K$,
we cast the $K$-dimensional multi-label classification as $K$ binary classifications. 
The predicted probability of each binary classification follow corresponding Dirichlet distribution.
Here, the Dirichlet distribution reduces to a Beta distribution.  
For generality, we still describe EDL for multi-label sewer defect classification via Dirichlet distribution.

Evidential generation module (EGM) is conducted to generate the defective ($+$) and non-defective ($-$) evidences $\mathbf{e}_K = \{ (e_{1}^{+}, e_{1}^{-}),  \dots, (e_{K}^{+}, e_{K}^{-}) \}$ of $K$ binary sewer defect classifications as follows:
\vspace{-1mm}
\begin{equation}
	\mathbf{e}_K = \phi (\mathbf{w}^{\mathsf{T}}\mathbf{f}^* + \mathbf{b}), 
	\vspace{-1mm}
\end{equation}
where $\mathbf{w} \in \mathbb{R}^{C \times 2K}$ and $\mathbf{b} \in \mathbb{R}^{2K \times 1}$ refer to the weight vector and bias, respectively. 
$\phi$ denotes a non-negative activation function which is ReLU empirically.
$\mathbf{f}^* \in \mathbb{R}^{C \times 1}$ is the feature vector of deep feature $\mathbf{f}$ pooled via global average operation.
Intuitively, we can exploit Eq.~\ref{equ:pro_opinion} (or Eq.~\ref{equ:pro_dir}) to 
derive the defect probability $\mathbf{p}_K = \{ (p_{1}^{+}, p_{1}^{-}), \dots, (p_{K}^{+}, p_{K}^{-}) \}$ based on the set of $K$ Dirichlet distributions $\{ Dir(\mathbf{p}_{k} | \boldsymbol{\alpha}_{k
})\}_{k=1}^{K} $, in which the Dirichlet parameter $\boldsymbol{\alpha}_{k} = {(\alpha_{k}^{+}, \alpha_{k}^{-})}$ is given by:
\vspace{-1mm}
\begin{equation}
		\alpha_{k}^{i} = e_{k}^{i} + a_{k}^{i}W,
\vspace{-1mm}
\end{equation}
where $i \in {\{+,-\}}$ from a set including defective and non-defective indicators.
$ a_{k}^{+}$, $a_{k}^{-}$ and $W$ are set $1/2$, $1/2$ and $2$, respectively.

\subsection{Expert base rate assignment}

Intuitively, the importance of defect classes are different in the practice, 
we introduce the expert knowledge to reassign the base rates of each defect class via expert base rate assignment (EBRA).
The realistic base rates based on expert knowledge would enhance the reliability of determination intuitively \cite{josang2016subjective}. 
Here, we utilize class-importance weights (CIW) as the expert knowledge, which is normalized by \cite{Haurum2021Sewer}. 
The procedure of EBRA for the $k$-th binary classification can be formulated as follows:
\vspace{-1mm}
\begin{equation}
		\hat{a}_{k}^{i} =  a_{k}^{i} + (-1)^{ \llbracket  i \neq + \rrbracket}(\sigma(\text{CIW}_{k}) - 1/2) ,
		\vspace{-1mm}
\end{equation}
where $\llbracket  \cdot \rrbracket$ is the indicator function which takes 1 when the argument is true and 0 otherwise. 
$\sigma$ denotes a sigmoid function, and $\text{CIW}_{k}$ is the class-importance weight of the $k$-th defect class.
Subsequently, the probability mass, uncertainty, belief mass, and Dirichlet parameter of $K$ binary classifications can be derived by formulations in Section~\ref{sec:edl} based on 
$\hat{\mathbf{a}}_{K}=\{(\hat{a}_{k}^{+}, \hat{a}_{k}^{-})\}_{k=1}^{K}$, 
termed as $\hat{\mathbf{p}}_{K} =\{(\hat{p}_{k}^{+}, \hat{p}_{k}^{-})\}_{k=1}^{K}$, 
$\hat{\mathbf{u}}_{K} =\{\hat{u}_{k}\}_{k=1}^{K}$,
$\hat{\mathbf{b}}_{K} =\{(\hat{b}_{k}^{+}, \hat{b}_{k}^{-})\}_{k=1}^{K} $ and 
$\boldsymbol{\hat{\alpha}}_{K} = \{(\hat{\alpha}_{k}^{+}, \hat{\alpha}_{k}^{-})\}_{k=1}^{K}$, respectively.

\subsection{Training and inference}

The training procedure of EDL is conducted based on the Type \uppercase\expandafter{\romannumeral2} Maximum Likelihood (Empirical Bayes) \cite{JAMIL2012Selection}.
We first obtain the evidences $\mathbf{e}_K$ from EGM, and then, 
convert multi-class label $\mathbf{y} \in \mathbb{H}^{K \times 1} $ as $K$ binary class labels $\mathbf{y}_{K} = \{(y_{1}^{+} , y_{1}^{-}), \dots, (y_{K}^{+} , y_{K}^{-})\}$,
where $\mathbb{H}$ denotes Hamming space.
The loss function of $k$-th binary classification can be formulated as a minimization of negative log-likelihood: 
\vspace{-1mm}
\begin{equation}
	\begin{aligned}
		\mathcal{L}_{k} &=-\log \left(\int \prod^{i \in {\{+,-\}}} \!\!\!\! (\hat{p}_{k}^{i})^{y^{i}_{k}} \frac{1}{B\left(\boldsymbol{\alpha}_i\right)} \!\!\!\! \prod^{i \in {\{+,-\}}} \!\!\!\! (\hat{p}_{k}^{i})^{\alpha_{k}-1} d \mathbf{\hat{p}}_{k}\right) \\
		&=\!\! \sum^{i \in {\{+,-\}}} \!\!\!\! y_{k}^{i} \left(\log ( \hat{S}_{k} ) -\log (\hat{\alpha}_{k}^{i})\right) \\
		&=\!\! \sum^{i \in {\{+,-\}}} \!\!\!\! y_{k}^{i} \left(\log ( \hat{S}_{k} ) -\log (\hat{e}_{k}^{i} + \hat{a}_{k}^{i} W)  \right),
	\end{aligned}
\end{equation}
where $\hat{S}_{k} = \sum^{i \in {\{+,-\}}} \hat{\alpha}_{k}^{i}$. 
Eventually, TMSDC for $K$ binary sewer defect classifications can be optimized as follows:
\vspace{-2mm}
\begin{equation*}
	\underset{ \theta_{F}, \theta_{G} }{\arg \min \ } \sum\nolimits_{k=1}^{K} \mathcal{L}_{k}.
\vspace{-1mm}
\end{equation*}
In the inference, we utilize maximum operation to obtain the uncertainty estimation $\hat{u}$ of sewer pipe image $\mathbf{x}$ as follows:
\vspace{-1mm}
\begin{equation}\label{equ:agg}
	\hat{u} = \underset{ k}{\max} ~ \hat{u}_{k}
\vspace{-1mm}
\end{equation}

\section{Experiments}
\label{sec:exp}

\subsection{Experimental setup}
\textbf{Dataset.} 
Sewer-ML \cite{Haurum2021Sewer} is a large-scale benchmark dataset, which focuses on the multi-label sewer defect classification task.
1.3 million sewer pipe images are collected over a nine year period, 
annotated with 17 defect classes and divided into 3 subsets in terms of training, validation, and testing.
The numbers of samples for 3 subsets are 1,040,129, 130,046, and 130,026, respectively. 
Since the annotations of testing set are not public, we conduct the experiments focused on the training and validation sets.

\textbf{Evaluation tasks and metrics.} 
To evaluate the performance of TMSDC on both known and unknown settings,
we conduct the evaluation tasks as follows: multi-label sewer defect classification $\mathcal{T}_{\text{MSDC}}$ and out-of-distribution (OOD) detection $\mathcal{T}_{\text{OOD}}$, separately.
Specifically, $\mathcal{T}_{\text{MSDC}}$ is utilized to demonstrate TMSDC capability of classifying the known sewer defect categories,
in which $\mathrm{F1_{Normal}}$ and $\mathrm{F2_{CIW}}$ are introduced as evaluation metrics following \cite{Haurum2021Sewer}.
In $\mathcal{T}_{\text{OOD}}$, we select the partial categories as the unknown defect categories, 
and the sewer samples with unknown defect labels are regarded as unknown samples.
It means that the samples with unknown defect labels would not be ``seen'' in the training phase.
The uncertainty estimation $\hat{u}$ of TMSDC is leveraged to distinguish the unknown and known samples.
The evaluation metrics for the OOD detection task are AUROC, AUPR, and FPR95,
where the unknown samples are defined as the positive cases.

\textbf{Implementation details.}
The experiments are carried out on a work station with NVIDIA Tesla A100 GPUs. 
The proposed method is implemented based on PyTorch deep learning framework.
The backbone of feature extractor ${F}$ is TResNet-L \cite{ridnik2021tresnet}.
In the training phase of $\mathcal{T}_{\text{MSDC}}$, the learnable parameters of TMSDC are trained from scratch via stochastic gradient descent (SGD) with a weight decay of 1e-4. 
The total training epochs are 90 and the initial learning rate is 1e-1, 
while the learning rate is reduced with the decay ratio of 0.1 after every 30 epochs. 
The batch size of training is 256.
The input images are scaled as $224 \times 224$, while random flip, jitter of pixel values (such as brightness, contrast, saturation, and hue) are used as data argumentation.
In $\mathcal{T}_{\text{OOD}}$, RB, OB, FS and OS are selected as unknown defect categories based on higher CIW,
since it is more valuable to verify the model performance of uncertainty estimation for serious unknown sewer defect samples.
For a fair comparison, we utilize a shared weight feature extractor for TMSDC and other OOD methods. 
We first train TResNet-L $F_{S}$ with a fully connected layer-based multi-label classifier $M$ via hyper-parameters in $\mathcal{T}_{\text{MSDC}}$.
Then, $M$ is replaced by EGM $G$, and the trained parameters of $F_{S}$ are fixed. $G$ is fine-tuned for 20 epochs with 1e-3 learning rate.
To simplify the description, the architectures of TMSDC and other OOD methods are denoted as $F_{S} \circ G$ and $F_{S} \circ M$, respectively.

\subsection{Comparisons with the state-of-the-art}

\begin{table}
	\centering
	\caption{Comparison of our method with the state-of-the-art methods of $\mathcal{T}_{\text{MSDC}}$ on Sewer-ML. 
		$\dag$ represents the two-stage-based method.
	}
	\label{tab:sota_msdc}
\resizebox{0.8\linewidth}{!}{  
	\begin{tabular}{ccc}
		\toprule
		\multirow{2}{*}{Method} & \multicolumn{2}{c}{Validation} \\
		& $\mathrm{F2_{CIW}} (\%) \uparrow$             & $\mathrm{F1_{Normal}} (\%) \uparrow$            \\
		\midrule
		Xie\dag \cite{Xie2019Automatic}                   &    48.57            &   91.08            \\
		\rowcolor{gray!10}
		Chen\dag \cite{Chen2018Intelligent}                  &    48.67            &   91.06            \\
		ResNet-101 \cite{He2016Deep}            &    53.26            &   79.55            \\
		\rowcolor{gray!10}
		KSSNet \cite{Wang2020Multi}                &  54.42              &   80.60            \\
		TResNet-L \cite{ridnik2021tresnet}            &    54.63            &   81.22            \\
		\rowcolor{gray!10}
		\textbf{TMSDC} (Ours)           &   54.54             &   81.15         \\
		\bottomrule
	\end{tabular}
}
\end{table}
\vspace{-1mm}

\textbf{Multi-label sewer defect classification $\mathcal{T}_{\text{MSDC}}$.}
We compare our method with 5 state-of-the-art methods lately reported on Sewer-ML,
which can be categorized into two-stage and end-to-end methods.
As shown in Tab.~\ref{tab:sota_msdc}, we observe that TMSDC achieve competitive performances of $17$ defect classes ($K=17$) in terms of $\mathrm{F2_{CIW}}$ and $\mathrm{F1_{Normal}}$.
It validates the acceptable performance of TMSDC for classifying known multi-label sewer defect samples.

\begin{table}
	\centering
	\caption{Comparison of our method with the competitive methods of $\mathcal{T}_{\text{OOD}}$ on Sewer-ML. 
	}
	\label{tab:sota_ood}
	\resizebox{\linewidth}{!}{  
	\begin{tabular}{ccccc}
		\toprule
		\multirow{2}{*}{Method} & \multirow{2}{*}{Arch} & \multicolumn{3}{c}{Validation} \\
		&                               & AUROC $\uparrow$     & AUPR  $\uparrow$   & FPR95 $\downarrow$     \\
		\midrule
		MaxLogit \cite{hendrycks2019scaling}            & $F_{S} \circ M$            & 65.19     & 77.40        & 83.03         \\
		\rowcolor{gray!10}
		JointEnergy \cite{Wang2021Can}            &         $F_{S} \circ M$    &   81.15      & 90.33         &  81.35   \\
		SLCS \cite{WANG2022Multi}                  &         $F_{S} \circ M$     &   81.97      & 91.11         &  77.14           \\
		\rowcolor{gray!10}
		\textbf{TMSDC} (Ours)                    & $F_{S} \circ G$  &  85.56  &   92.23   &   55.83   \\
		\bottomrule
	\end{tabular}
}
\end{table}

\textbf{OOD detection $\mathcal{T}_{\text{OOD}}$.}
We compare our method against competitive OOD detection methods of multi-label classification, 
in which TMSDC demonstrates state-of-the-art performance, as reported in Tab.~\ref{tab:sota_ood}.
Here, the performances of $F_{S} \circ M$ for 13 defect classes ($K= 13$) are close to $F_{S} \circ G$.
$\mathrm{F2_{CIW}}$ of two architectures are 54.67\% and 54.61\%, 
while $\mathrm{F1_{Normal}}$ of them are 84.34\% and 84.27\%.
These facts verify that TMSDC achieves the reliable results on unknown uncertainty estimation with a tolerable decrease of classification performance.

\subsection{Ablation studies}

\begin{table}
	\centering
	\caption{Ablation analysis of TMSDC on Sewer-ML. \checkmark denotes TMSDC with EBRA.
	}
	\label{tab:abla}
	\resizebox{0.9\linewidth}{!}{  
		\begin{tabular}{ccccc}
			\toprule
			\multirow{2}{*}{w/ EBRA} & \multirow{2}{*}{Aggregation} & \multicolumn{3}{c}{Validation} \\
			&                              & AUROC $\uparrow$    & AUPR  $\uparrow$  & FPR95 $\downarrow$   \\
			\midrule
			\rowcolor{gray!10}
			& Max                          &  82.78         &   91.30      &   77.84      \\
			\checkmark & Max                          &      85.56  &   92.23   &   55.83         \\
			\midrule
			\rowcolor{gray!10}
			\checkmark & Sum                          &    82.24         &   91.09      &   80.56         \\
			\checkmark & Top-5                          &  86.11  &   92.69   &   67.21        \\
			\rowcolor{gray!10}
			\checkmark & Max                          &  85.56  &   92.23   &   55.83         \\ 
			\bottomrule
		\end{tabular}
	}
\end{table}

\textbf{Effectiveness of expert base rate assignment.} 
To clarify the effectiveness of EBRA, we train TMSDC with a average base rate whose performance is reported in Tab.~\ref{tab:abla}.
It can be seen that EBRA improves the capability of distinguishing unknown defect categories for TMSDC obviously,
validating the effectiveness of EBRA.

\textbf{Impact of different aggregation method in Eq.~\ref{equ:agg}.} 
To explore the impact of different aggregation method, 
we utilize three aggregation methods for uncertainty estimation alternately including summation (Sum),  top-5 (Top-5) and maximum (Max) operations.
Top-5 means that the summation of the fifth highest uncertainty scores. 
As listed in Tab.~\ref{tab:abla}, 
the summation operation might worsen the discriminability of uncertainty estimation, resulting in the lower performance.
The top-5 operation outperforms the maximum operation in terms of AUROC and AUPR,
in which the maximum operation achieves the lowest false positive rate of unknown samples where the true positive rate of known samples is at 95\%.


\section{Conclusion}
\label{sec:conclusion}
In this paper, we propose a trustworthy multi-label sewer defect classification (TMSDC) method, 
which can quantify the uncertainty of sewer defect prediction via evidential deep learning.
Meanwhile, a novel expert base rate assignment (EBRA) is proposed to introduce the expert knowledge for describing reliable evidences in practical situations.
Experimental results demonstrate that TMSDC is effective and achieves the superior capability of uncertainty estimation on the latest benchmark.


\bibliographystyle{IEEEbib}
\bibliography{20220930_icassp_01}

\end{document}